\documentclass[final]{csri24}

\usepackage[top=1.355in,
			bottom=1.355in,
			left=1.5in,
			right=1.5in,
			heightrounded]{geometry}
\usepackage{amsfonts,
			amsmath,
			amssymb,
			graphicx,
			subfigure,
			url,
            color}
\usepackage[normalem]{ulem}
\graphicspath{{figures/}{../figures/}}

\usepackage{boldtensors}
\usepackage{booktabs}
\usepackage{array}

\newcommand{\Pe}{\mathrm{Pe}}

\title{Hybrid coupling with numerics-informed neural networks and the overlapping Schwarz alternating method}

\author{George Chumbipuma\thanks{Rice University, gc51@rice.edu} \and Irina Tezaur\thanks{Sandia National Laboratories, ikalash@sandia.gov} \and Alejandro Diaz\thanks{Sandia National Laboratories, andiaz@sandia.gov} \and Beatrice Riviere\thanks{Rice University, riviere@rice.edu}} 

\begin{document}

\maketitle

\begin{abstract}
We develop a hybrid modeling framework for coupling pre-trained numerics-informed neural networks (NINNs) with classical full order models (FOMs) using the overlapping Schwarz alternating method.  We consider the two-dimensional advection-diffusion equation in the advection-dominated, P\'{e}clet-number \(10^6\) regime. We first demonstrate that, unlike the corresponding physics-informed neural network (PINN), a monolithic NINN can be accurately trained on our model problem without domain decomposition. We then employ overlapping multiplicative Schwarz as a \textit{deployment} mechanism for coupling a pre-trained, subdomain-local NINN with a neighboring FOM, with the NINN weights held fixed throughout the Schwarz iteration. We consider two training approaches for the subdomain-local NINNs: a top-down approach, in which boundary data are obtained from a coupled Schwarz solve on the full domain with a FOM on each subdomain (FOM-FOM Schwarz), and a bottom-up approach, in which boundary traces are generated synthetically on the NINN subdomain without requiring any full-domain solves. The resulting hybrid NINN-FOM solutions agree closely with the corresponding FOM-FOM Schwarz solutions, with the top-down and bottom-up training approaches yielding comparable accuracy.

\end{abstract}

\section{Introduction}
\label{sec:intro}


It is well-known that the advection-diffusion equation, a canonical model problem for transport phenomena arising in fluid dynamics, poses significant numerical challenges in the advection-dominated, high P\'{e}clet-number regime. Solutions in this regime often exhibit sharp boundary layers that, if inadequately resolved, can lead to spurious oscillations and loss of numerical accuracy \cite{Kalashnikova2009}. These sharp, transport-dominated features also give rise to solution families with slowly decaying Kolmogorov \(n\)-width, making accurate low-dimensional approximation challenging~\cite{mojgani2023kolmogorov}. The spatially localized nature of these features, together with the disparate length scales present in the solution, naturally motivates the use of different local models in different regions of the domain. For example, it is natural to employ a classical solver in regions containing the sharpest gradients, while assigning a reusable learned model elsewhere in the domain
 \cite{barnett2022schwarz,moore2024opinfschwarz,tezaur2025hybrid}.
In a repeated-query setting, this sort of domain decomposition-based model assignment and coupling has the potential to reduce online computational cost while retaining the accuracy of classical models in regions where it is most needed.

The Schwarz alternating method constructs a global solution by iteratively solving local problems on overlapping subdomains, with the solution from each subdomain providing updated  boundary conditions for neighboring subdomains at each iteration~\cite{schwarz1870,lions1988schwarz,gander2008schwarz}.  Importantly, the local models need not be identical, making the Schwarz framework naturally suited to heterogeneous model coupling. Previous work has demonstrated this capability by coupling disparate meshes and time integrators in solid mechanics~\cite{mota2017schwarz,mota2022schwarz}, as well as by coupling projection-based reduced-order models (ROMs), operator-inference ROMs, and physics-informed neural networks (PINNs) with full order models (FOMs)~\cite{barnett2022schwarz,moore2024opinfschwarz,tezaur2025hybrid,snyder2023schwarz}. In this work, we extend our previously-developed overlapping  heterogeneous Schwarz coupling framework to an emerging class of data-driven models termed ``numerics-informed neural networks" NINNs~\cite{celaya2024ninn}, demonstrating the coupling of pre-trained, subdomain-local NINNs with subdomain-local FOMs for the steady two-dimensional (2D) advection-diffusion equation.

The present work builds on the Schwarz-based PINN coupling framework of Snyder et al.~\cite{snyder2023schwarz}, developed for the one-dimensional (1D) steady advection-diffusion equation. Their primary focus was the use of domain decomposition and Schwarz iteration to facilitate PINN training: subdomain-local PINNs were trained concurrently with the Schwarz iteration as interface data were exchanged between neighboring subdomains. Although this approach did not substantially improve PINN trainability at high P\'{e}clet numbers, coupling a PINN with a FOM on the outflow subdomain enabled convergence at a P\'{e}clet number of \(10^6\). The complementary problem of using Schwarz to couple pre-trained learned models, without further training during the Schwarz iteration, was left for future investigation.

In recent years, NINNs have emerged as a promising alternative to PINNs, demonstrating improved accuracy and greater ease of training in a number of settings \cite{celaya2024ninn, celaya2025dgninn}.  The key distinction between PINNs and NINNs is that, whereas PINNs enforce the strong form partial differential equation (PDE) residual in their loss function being minimized, NINNs instead construct the loss from a discretized form of the PDE obtained using a classical numerical method, e.g., finite differences \cite{celaya2024ninn} or the discontinuous Galerkin (DG) method \cite{celaya2025dgninn}.  By embedding a prescribed numerical discretization into the training objective, a NINN learns the coefficients associated with that discretization rather than directly approximating a continuous solution satisfying the strong form PDE, thereby simplifying the learning problem.

In this work, we first demonstrate that a monolithic NINN with strong Dirichlet boundary condition (SDBC) enforcement can accurately solve a 2D advection-diffusion problem at the ultra-high P\'{e}clet number of \(10^6\) with minimal training, a regime in which a corresponding PINN fails to achieve an accurate solution. We then investigate the use of the Schwarz alternating method to couple a pre-trained, subdomain-local NINN with a subdomain-local FOM, with no additional NINN training performed during the Schwarz iteration. We explore two training approaches for the NINN: a \textit{top-down} approach and a \textit{bottom-up} approach. In the top-down approach, training interface data are collected from a FOM-FOM Schwarz coupling and subsequently used to train the NINN, yielding traces that are representative of the coupled problem but requiring coupled FOM simulations during the offline stage. In contrast, in the bottom-up approach, the interface traces are instead generated synthetically from a prescribed family of functions, avoiding the need for FOM-FOM Schwarz training simulations but requiring that the synthetic traces adequately represent the interface conditions encountered during deployment. We assess the proposed hybrid NINN-FOM coupling on the steady advection-diffusion problem in the high P\'{e}clet-number regime. The pre-trained NINN can be successfully coupled with a FOM through the Schwarz alternating method, yielding solutions in close agreement with the corresponding FOM-FOM Schwarz solutions. Moreover, the top-down and bottom-up training approaches yield comparable coupling accuracy, demonstrating that accurate hybrid coupling can be achieved without requiring FOM-FOM Schwarz simulations to generate the NINN training data.

The remainder of this paper is organized as follows.  In Section \ref{sec:pdes}, we present the 2D steady advection-diffusion problem considered herein.  
Section~\ref{sec:ninns} reviews the NINN formulation, including its Pocket U-Net architecture and its training objective, and discusses key differences between the NINN and PINN approaches.
Section~\ref{sec:schwarz} describes the overlapping multiplicative Schwarz alternating method for heterogeneous NINN-FOM coupling, and the top-down and
bottom-up training approaches used to pre-train the NINNs used in these couplings.
  Section~\ref{sec:results} presents the numerical results, beginning with an assessment of monolithic NINN trainability in the advection-dominated regime at the ultra-high P\'{e}clet number of $10^
6$, followed by demonstrations of pre-trained NINN-FOM Schwarz coupling, both for a steady advection-diffusion boundary value problem. Section~\ref{sec:conc} summarizes our findings, and discusses conclusions as well as future work.



\section{Model problem} \label{sec:pdes}

\begin{figure}[ht!]
\centering
\includegraphics[width=0.4\textwidth]{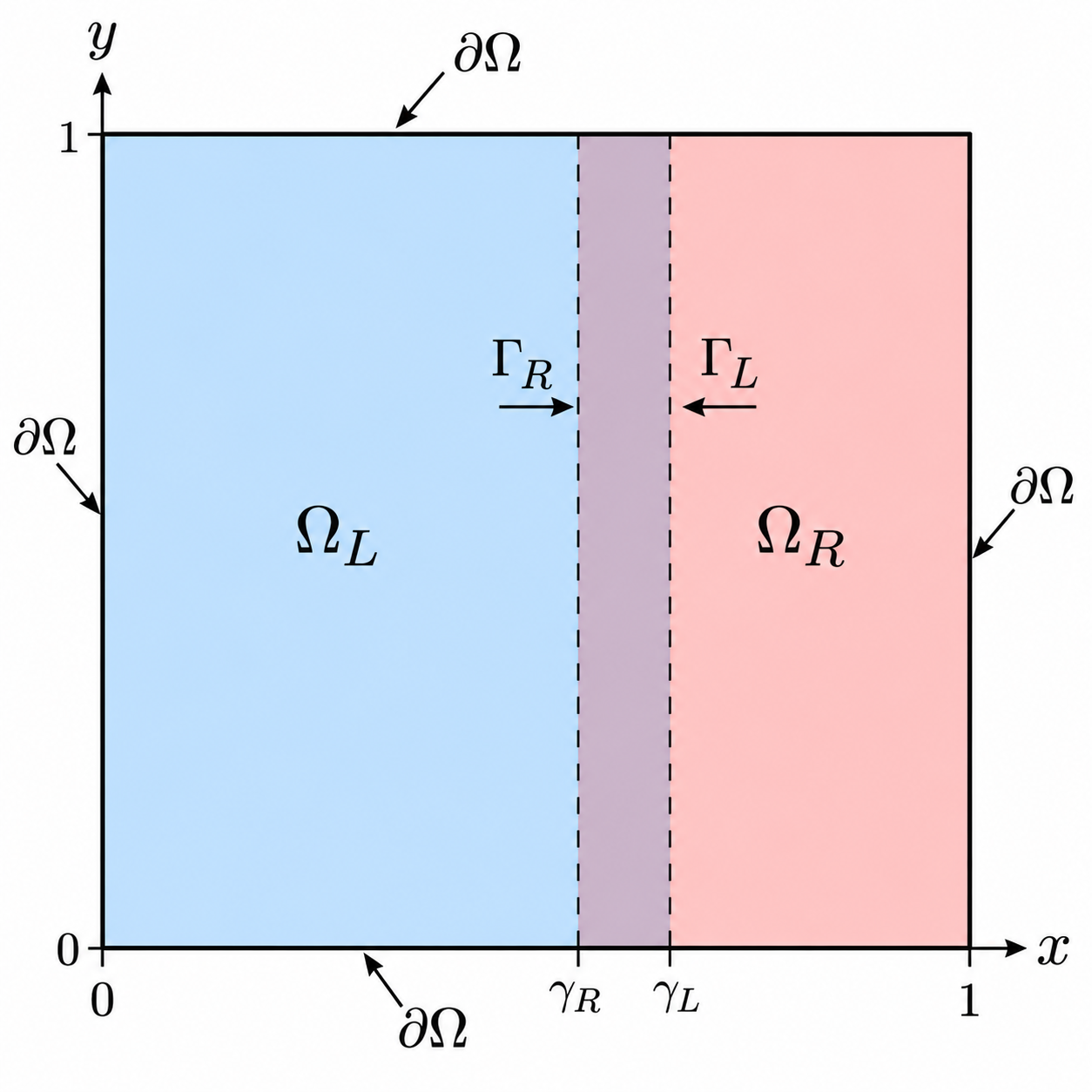}
\caption{Spatial domain $\Omega:=(0,1) \times (0,1)$ and its decomposition into two overlapping subdomains $\Omega_L := (0,\gamma_L) \times (0,1)$ and $\Omega_R := (\gamma_R, 1) \times (0,1)$.}
\label{fig:dd}
\end{figure}

We consider herein the 2D steady advection-diffusion equation: 
\begin{equation}
\label{eq:model}
\mathcal{L}u=f
\quad\text{in }\Omega:=(0,1)\times (0,1),
\qquad
u=0\quad\text{on }\partial\Omega,
\end{equation}
where 
\begin{equation} \label{eq:operator}
    \mathcal{L}u:=~\beta\cdot\nabla u-\nu\Delta u.
\end{equation}
Here, $~\beta \in \mathbb{R}^2$ is the (constant) advection-field, $\nu \in \mathbb{R}$ is the viscosity, and $f\in L^2(\Omega)$ is the source function.  The P\'{e}clet number is defined as $Pe_L:=|~\beta|L/\nu = |~\beta|/\nu$, since $L$, the length scale associated with the geometry, is 1. Figure \ref{fig:dd} depicts the spatial domain $\Omega$ with boundary $\partial\Omega$, as well as its decomposition into two overlapping subdomains, $\Omega_L:=(0,\gamma_L)\times (0,1)$ and $\Omega_R:=(\gamma_R,1)\times(0,1)$, where $0< \gamma_R < \gamma_L < 1$, which will be referenced in later sections.      
We will restrict attention herein to the case of a horizontal advection field, so that $~\beta=(1,0)$.  In this case, the solution has an outflow layer of width \(O(\varepsilon)\) at \(x=1\)
and characteristic layers of width \(O(\sqrt{\varepsilon})\) at \(y=0,1\),
where \(\varepsilon:=\nu/|~\beta|\).

\section{Numerics-Informed Neural Networks (NINNs)} \label{sec:ninns}


Numerics-informed neural networks (NINNs) are a class of physics-informed neural
networks in which a numerical discretization of the governing PDE is embedded directly into the training
objective. To clarify the distinction between NINNs and conventional
PINNs, consider the steady
advection-diffusion problem~\eqref{eq:model} with differential
operator $\mathcal{L}$ defined in~\eqref{eq:operator}.  For simplicity and because it produced superior results, we consider strong enforcement of the Dirichlet boundary conditions in both formulations. In each case, a boundary-conforming ansatz (lifting) is applied to the network output such that the resulting solution satisfies the prescribed Dirichlet data exactly \cite{snyder2023schwarz}. Consequently, no boundary loss term is required, so we omit it from our presentation of the PINN and NINN formulations. 

\begin{table}[ht!]
\centering
\caption{Comparison of the primary PINN and NINN formulations used in this work.}
\label{tab:pinn_ninn_comparison}
\renewcommand{\arraystretch}{1.35}

\begin{tabular}{
    >{\raggedright\arraybackslash}p{0.28\textwidth}
    >{\raggedright\arraybackslash}p{0.30\textwidth}
    >{\raggedright\arraybackslash}p{0.34\textwidth}
}
\toprule
& \textbf{PINN} & \textbf{NINN} \\
\midrule

\textbf{Primary architecture}
& Coordinate-based MLP
& Pocket U-Net (CNN) \\

\midrule
\textbf{Input/output representation}
& Coordinates $~x \rightarrow u_\theta(~x)$
& Mesh-based fields $~X \rightarrow \mathbf{u}_\theta$ \\

\midrule
\textbf{Activation function}
& Nonlinear $\tanh$
& Identity in convolutional blocks \\

\midrule
\textbf{Mesh dependence of architecture}
& Not tied to a computational mesh
& CNN operates on structured mesh arrays \\

\midrule
\textbf{Physics-based loss function}
& $\left\|\mathcal{L}u_\theta-f\right\|^2$
& $\left\|~A_h~u_\theta-~b_h\right\|_2^2$ \\

\midrule
\textbf{PDE derivatives}
& Automatic differentiation
& Finite differences \\

\midrule
\textbf{Underlying discretization}
& ---
& 2nd-order upwind advection; centered finite difference diffusion;
  1st-order inflow closure \\

\bottomrule
\end{tabular}
\end{table}

The primary differences between a typical PINN and a NINN architecture are summarized in Table \ref{tab:pinn_ninn_comparison}.  A conventional PINN \cite{raissi2019pinn} approximates the solution $u$ to \eqref{eq:model} as a continuous function $u_{\theta}(~x)$, typically parameterized by a multilayer perceptron (MLP) network.  
Here, and in what follows, $\theta$ denotes a set of trainable parameters, comprised of a set of weights and biases, of a NN.  
Given a set of $N_c \in \mathbb{N}^+$ collocation points $\{~x_i\}_{i=1}^{N_c} \in \Omega$ and assuming no data loss term, the PINN is trained by minimizing   
\begin{equation} \label{eq:pinn}
    J_{\text{PINN}}(\theta) = \frac{1}{N_c} \sum_{i=1}^{N_c} || \mathcal{L}u_{\theta}(~x_i) - f(~x_i)||^2.
\end{equation}
The objective function \eqref{eq:pinn} is typically minimized using an Adam optimizer \cite{kingma2015adam}, with the derivatives in $\mathcal{L}u_{\theta}$ evaluated using automatic differentiation.  Thus, the PINN maps a spatial coordinate $~x$ to an approximation of the solution at that coordinate, and is trained to approximately satisfy the continuous strong form PDE at a collection of collocation points.  

In contrast to a traditional PINN, a NINN begins with a numerical discretization of the governing PDE, which, in the case of our steady linear problem \eqref{eq:model}, gives rise to a linear discrete system of the form 
\begin{equation} \label{eq:discrete}
    ~A_h ~u_h = ~b_h,
\end{equation}
where $~A_h \in \mathbb{R}^{N_h \times N_h}$ is a matrix representing the discretization stencil, $~u_h \in \mathbb{R}^{N_h}$ is the discretized solution vector, and $~b_h \in \mathbb{R}^{N_h}$ contains the discretized source term $f$ and contributions from the prescribed Dirichlet data, with $N_h \in \mathbb{N}^+$.  
In the present work, $~A_h$ is constructed using a finite difference discretization consisting of a second-order upwind approximation of the advective term and a second-order approximation of the diffusive term, with a first-order upwind closure at the first interior node adjacent to the inflow boundary.  In a NINN, the discrete solution is represented as the nodal field
\begin{equation}
    ~u_{\theta} = N_{\theta} (~X),
\end{equation}
where $~X$ is a mesh-based input tensor.  The NINN parameters are then determined by minimizing the residual of the discrete system 
\eqref{eq:discrete}, given by 
\begin{equation} \label{eq:ninn_loss}
    J_{\text{NINN}}(\theta) = \frac{1}{N_h} || ~A_h ~u_{\theta} - ~b_h||_2^2.
\end{equation}
Similar to a PINN, the NINN loss \eqref{eq:ninn_loss} is typically optimized using the Adam algorithm.  While it is common to include in the PINN objective function \eqref{eq:pinn} a data misfit term, known as the data loss, to facilitate PINN training and improve accuracy, NINNs were developed as \textit{fully unsupervised} PDE solvers: the training objective is constructed from the residual of the numerical discretization and does \textit{not} require labeled solution data.  In the numerical results presented in Section \ref{sec:results}, neither the PINN nor the NINN considered uses labeled data.

\begin{figure}[ht!]
\centering
\includegraphics[width=\textwidth]{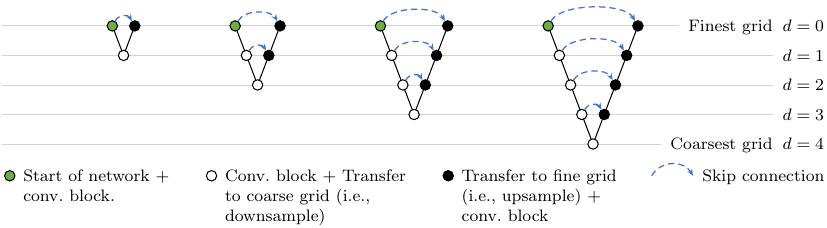}
\caption{Pocket U-Net encoder-decoder maps of depths 1--4,
with skip connections and channel width fixed across levels.
Reproduced from Celaya et al.~\cite{celaya2024ninn}.
The NINN in this paper uses depth \(3\), corresponding to three
max-pooling stages.}
\label{fig:pocket-unet}
\end{figure}

In addition to utilizing a different loss function, another key distinction between a PINN and a NINN is that, whereas a PINN is typically based on a coordinate-based MLP\footnote{In our numerical results comparing PINNs and NINNs, we also consider a PINN with a Pocket U-Net architecture in an effort to make the PINN vs. NINN comparison more consistent.  Please see Section \ref{sec:results} for details.} that maps a spatial coordinate $~x$ to the scalar approximation $u_{\theta}(~x)$, our NINN employs a Pocket U-Net architecture of Celaya et al. \cite{celaya2022pocketnet}, a convolutional encoder-decoder NN with skip connections (Figure \ref{fig:pocket-unet}).  The encoder uses max-pooling to progressively coarsen the spatial representation and increase the effective receptive field, while skip connections transfer fine-scale information to the decoder. This architecture is particularly well-suited to the finite difference NINN considered here because both the network inputs and outputs are represented as fields on a structured mesh, allowing convolutional operations to be applied directly to the mesh-based data, while the encoder-decoder structure enables information to be propagated across multiple spatial scales. The Pocket U-Net maintains a fixed number of feature channels across levels, providing a relatively compact Convolutional Neural Network (CNN) architecture. A limitation of this construction is its reliance on the regular array structure required by standard convolutional layers; consequently, the present NINN formulation is naturally restricted to structured grids. Extension to general unstructured meshes would require a different network representation or architecture; the interested reader is referred to \cite{celaya2025dgninn} for more details on how this can be done.

A final difference between the PINN and a NINN that is worth highlighting is the fact that the two networks utilize different activation functions.  Whereas a coordinate PINN employs a nonlinear activation function, in our case a $\tanh$ activation, the convolutional blocks of the Pocket U-Net NINN use identity functions, following the approach in \cite{celaya2022pocketnet, celaya2025dgninn}.  Because the underlying advection-diffusion problem and its finite difference discretization are linear, the Pocket U-Net follows \cite{celaya2022pocketnet, celaya2025dgninn} in using identity rather than nonlinear activation functions within its convolutional blocks. We note, however, that the complete NINN mapping is not strictly linear due to the presence of max-pooling operations.  Moreover, it is possible to utilize nonlinear activations within the NINN architecture, and we hypothesize that this may be needed when tackling nonlinear problems.

One key motivation for the NINN formulation is that it can simplify the learning problem relative to a PINN. A PINN must learn a continuous approximation whose spatial derivatives, evaluated through automatic differentiation, satisfy the governing PDE. This can lead to a difficult optimization problem when the solution contains sharp spatial features, since the network must accurately represent both the solution and the derivatives entering the PDE residual. A NINN instead embeds a prescribed numerical discretization into the training objective. Spatial differentiation is therefore represented by the discrete operator $~A_h$, and the network is trained to produce nodal values that satisfy the resulting algebraic equations. 
Effectively, a NINN is tasked with learning the coefficients of a prescribed numerical discretization rather than a continuous solution whose derivatives must satisfy the strong form PDE. This can substantially simplify the optimization problem, as demonstrated numerically in Section \ref{sec:results_mono}.   A clear downside of the NINN approach is that it requires an underlying mesh, whereas PINNs are a type of mesh-free method that requires only a set of collocation points at which the solution is evaluated.



\section{The overlapping Schwarz alternating method for NINN-FOM coupling} \label{sec:schwarz}

The primary objective of this work is to extend the overlapping Schwarz alternating method \cite{mota2017schwarz, mota2022schwarz, Wentland2025, snyder2023schwarz} to the coupling of pre-trained subdomain-local NINNs with subdomain-local FOMs. The motivation for this hybrid formulation is to leverage the computational efficiency of a pre-trained NINN while retaining the robustness and accuracy of a FOM in regions where the solution is particularly challenging to approximate. For the advection-dominated problems considered here, such difficulties arise primarily in the vicinity of sharp boundary layers. We therefore seek to localize the more expensive FOM solve to a subdomain containing the boundary layer region, while replacing the FOM by a pre-trained NINN over the remainder of the computational domain. The overlapping Schwarz method provides a natural framework for coupling these heterogeneous subdomain-local models through the iterative exchange of Dirichlet data on their artificial interfaces.

Toward this effect, suppose our underlying spatial domain $\Omega$ is decomposed into two overlapping subdomains $\Omega_L:=(0, \gamma_L)\times (0,1)$ and $\Omega_R = (\gamma_R, 1) \times (0,1)$ with $0 < \gamma_R < \gamma_L < 1$, as shown in Figure \ref{fig:dd}. Define the receiving artificial interfaces by
$\Gamma_L:=\partial\Omega_L\cap\Omega_R=\{\gamma_L\}\times(0,1)$ and
$\Gamma_R:=\partial\Omega_R\cap\Omega_L=\{\gamma_R\}\times(0,1)$.
For the steady advection-diffusion problem \eqref{eq:model}, the classical multiplicative Schwarz iteration with a FOM on each subdomain takes the form: 
\begin{equation} \label{eq:schwarz_iter}
\left\{\begin{aligned}
\mathcal{L}u_L^{(k+1)} &= f_L
&& \text{in } \Omega_L, \\
u_L^{(k+1)} &= 0
&& \text{on } \partial\Omega \cap \bar{\Omega}_L, \\
u_L^{(k+1)} &= u_R^{(k)}
&& \text{on } \Gamma_L, \\[1mm]
\end{aligned}\right. \qquad
\left\{\begin{aligned}
\mathcal{L}u_R^{(k+1)} &= f_R
&& \text{in } \Omega_R, \\
u_R^{(k+1)} &= 0
&& \text{on } \partial\Omega \cap \bar{\Omega}_R, \\
u_R^{(k+1)} &= u_L^{(k+1)}
&& \text{on } \Gamma_R,
\end{aligned}\right.
\end{equation}
for Schwarz iterations $k=0,1,...$, where $u_L^{(k)}$ and $u_R^{(k)}$ denote the solutions in $\Omega_L$ and $\Omega_R$, respectively, at the $k^{th}$ Schwarz iterations, $f_L$ and $f_R$ denote the source terms in $\Omega_L$ and $\Omega_R$, respectively,  and $\bar{\Omega}$ denotes the closure of $\Omega$. We initialize $u_L^{(0)}=u_R^{(0)}=0$ to commence the iteration. The Schwarz iteration \eqref{eq:schwarz_iter} continues until the differences in the exchanged interface traces between successive iterations are sufficiently small.  \\

\noindent \textit{Remark 1.} As shown in \cite{mota2017schwarz, mota2022schwarz, tezaur2025hybrid}, it is possible to use Schwarz to couple subdomains discretized using non-conformal meshes. In this case, transfer operators are needed to map the neighboring subdomain solution to the receiving artificial interface. In the computations reported here, the two local meshes use the same $y$-grid; however, each Schwarz interface ($\gamma_L$ and $\gamma_R$) lies between $x$-grid lines of the neighboring subdomain mesh.  The neighboring subdomain solution is therefore interpolated in the $x$-direction to provide the Dirichlet data at the receiving interface nodes.
For simplicity, we omit the explicit transfer operators from \eqref{eq:schwarz_iter} and \eqref{eq:schwarz_iter_ninn}.  

\subsection{Coupled NINN-FOM Schwarz formulation} \label{sec:ninn-fom_schwarz}


In extending \eqref{eq:schwarz_iter} to the coupling of subdomain-local NINNs, suppose without loss of generality that we wish to couple a pre-trained NINN in $\Omega_L$ to a finite difference FOM in $\Omega_R$.  In order to apply the Schwarz alternating method using a pre-trained NINN as a subdomain solver, the NINN must be able to accept the Dirichlet data prescribed on the artificial Schwarz interface as an input and predict the corresponding solution throughout the subdomain.  Toward this effect, we begin by pre-training offline a subdomain-local NINN in $\Omega_L$ over a range of interface boundary conditions on $\Gamma_L$ and forcing terms in $\Omega_L$, so that the NINN can be evaluated online for the changing Dirichlet data encountered during the Schwarz iteration.  Specifically, we consider a subdomain-local NINN of the form 
\begin{equation} \label{eq:N_theta}
    N_{\theta}: (f_L, g_{\Gamma}) \mapsto ~u_{\theta, L},
\end{equation}
where $\theta$ denotes the trainable NINN parameters.
The inputs to \eqref{eq:N_theta} include the local source $f_L$ and the Dirichlet data $g_{\Gamma}$ prescribed on the artificial interface $\Gamma_L$, and the output is the predicted solution in $\Omega_L$, denoted by $ ~u_{\theta, L}$.  For a given $f_L$ and $g_{\Gamma}$, the corresponding FOM solution in $\Omega_L$, denoted by $~u_{h,L}$, satisfies
\begin{equation}
    ~A_{h,L}~u_{h,L} = ~b_{h,L}(f_L, g_{\Gamma}),
\end{equation}
so that the objective of offline NINN training is to learn an approximation to this local FOM solution map: 
\begin{equation}
    N_{\theta}(f_L, g_{\Gamma}) \approx ~A_{h,L}^{-1} ~b_{h,L}(f_L, g_{\Gamma}).
\end{equation}

In applying Schwarz to our NINN-FOM coupling scenario, the pre-trained NINN \eqref{eq:N_theta} replaces the FOM solve on $\Omega_L$, so that the Schwarz iteration takes the form 
\begin{equation} \label{eq:schwarz_iter_ninn}
\left\{\begin{aligned}
~u_{\theta^*, L}^{(k+1)} &= N_{\theta^*}(f_L, u_R^{(k)}|_{\Gamma_L})&& \text{in } \Omega_L,\\[1mm]
\end{aligned}\right. \qquad
\left\{\begin{aligned}
\mathcal{L}u_R^{(k+1)} &= f_R
&& \text{in } \Omega_R, \\
u_R^{(k+1)} &= 0
&& \text{on } \partial\Omega \cap \bar{\Omega}_R, \\
u_R^{(k+1)} &= u_{\theta^*, L}^{(k+1)}
&& \text{on } \Gamma_R,
\end{aligned}\right.
\end{equation}
for $k=0, 1, ...$, where $\theta^*$ denotes the NINN parameters obtained during offline training.  
In particular, at Schwarz iteration $k+1$, the NINN receives the interface trace from the previous right subdomain solution $u_R$ and predicts the updated solution on $\Omega_L$.  The resulting left subdomain solution then provides the Dirichlet data for the FOM solve on $\Omega_R$.  The iteration \eqref{eq:schwarz_iter_ninn} is continued until the exchanged interface data converge.  To monitor convergence, we collect Dirichlet traces exchanged between the two subdomains at Schwarz iteration $k$ into the interface vector 
\begin{equation}
    ~d^{(k)} = \left( \begin{array}{c} 
    ~u_{\theta^*,L}^{(k)}|_{\Gamma_R} \\
     ~u_{R}^{(k)}|_{\Gamma_L}
    \end{array}\right).  
\end{equation}
Convergence is declared when the change in the interface data between successive Schwarz iterations satisfies 
\begin{equation} \label{eq:schwarz_conv}
    ||~d^{(k+1)} - ~d^{(k)}||_{\infty} \leq \tau_a + \tau_r \max \left\{||~d^{(k)}||_{\infty}, ||~d^{(k+1)}||_{\infty} \right\},
\end{equation}
where $\tau_a>0$ and $\tau_r>0$ are prescribed absolute and relative Schwarz convergence tolerances, respectively.

It is important to emphasize that the NINN parameters $\theta^*$ are \textit{fixed} during the online Schwarz iteration \eqref{eq:schwarz_iter_ninn}; only the interface data supplied to the network change from one Schwarz iteration to the next.    This distinguishes the present approach from the earlier PINN-coupling framework of Snyder et al. \cite{snyder2023schwarz}, in which the subdomain PINNs are trained as part of the Schwarz iteration. In the present work, the NINN is instead trained offline to act as a reusable subdomain solver for varying interface boundary conditions.

\subsection{Offline training strategies} \label{sec:ninn_training}

The pre-trained NINN utilized within the Schwarz formulation described in Section \ref{sec:ninn-fom_schwarz} must be able to predict the local solution for the range of Dirichlet data that may be encountered on the artificial Schwarz interface. We consider two approaches for generating these interface boundary conditions for offline training: a top-down approach, in which representative interface data are obtained from FOM-FOM Schwarz simulations, and a bottom-up approach, in which the interface data are generated synthetically.  
These data consist of spatial interface traces.  
The two training strategies are summarized succinctly below.

\subsubsection{Top-down training} \label{sec:top-down}

In the top-down training approach, FOM-FOM Schwarz simulations are first performed for a collection of representative training problems. These training problems correspond to different source terms $f$ sampled from a prescribed family of functions.  For each training problem, the Dirichlet traces generated on the artificial interface during the Schwarz iteration are saved, and subsequently used as boundary condition inputs for training the NINN.  In this way, the NINN is trained using interface conditions that arise naturally during representative Schwarz solves. The disadvantage of the top-down training approach is that it requires performing coupled FOM-FOM simulations to generate the training data.

\subsubsection{Bottom-up training} \label{sec:bottom-up}

In the bottom-up training approach, the interface Dirichlet data are  generated synthetically on the artificial interface of the subdomain to be replaced by the NINN, $\Gamma_L$, without performing any FOM-FOM Schwarz simulations on the full domain $\Omega$. The traces are sampled from a prescribed family of functions chosen to represent a sufficiently broad range of possible interface conditions. These synthetic traces are then imposed as boundary conditions when training the local NINN. The main advantage of the bottom-up training approach is that it avoids the need for coupled FOM simulations to generate the training data; however, its effectiveness depends on whether the synthetic training family adequately represents the interface conditions encountered during online Schwarz coupling.

In this particular work, the synthetic interface traces on the NINN Schwarz boundary $\Gamma_L$ are generated using a truncated sine expansion, 
\begin{equation} \label{eq:sine}
\widehat{\gamma}(y)
=
\sum_{m=1}^{M}
\frac{\xi_m}{m^{3/2}}
\sin(m\pi y),
\end{equation}
where the coefficients $\xi_m\in \mathbb{R}$ are randomly sampled, and $M \in \mathbb{N}^+$ is a pre-selected integer. The decay of the coefficients favors smooth traces, while the sine basis ensures that the interface data are compatible with the homogeneous Dirichlet conditions at $y=0$ and $y=1$. Each trace is subsequently scaled to vary its amplitude. The specific sampling distributions and training set sizes used in the numerical experiments are given in Section \ref{sec:results}.

\section{Numerical results} \label{sec:results}


We now assess our NINN and Schwarz methodologies on the test cases described in Section \ref{sec:pdes}.  
We report results from two experiments: (i) monolithic NINN training (Section \ref{sec:results_mono}), and (ii) pre-trained NINN-FOM Schwarz coupling (Section \ref{sec:results_steady}).  

Except for the P\'{e}clet-number sweep reported in Table \ref{tab:Pe_sweep}, all testing is performed at $Pe_L=10^6$. For the domain decomposed study, we set $\gamma_L = 0.95$ and $\gamma_R = 0.65$ in Figure \ref{fig:dd}, so that $\Omega_L = (0,0.95) \times (0,1)$ and $\Omega_R = (0.65,1) \times (0,1)$, yielding an overlap region having width 0.3.  
In the NINN-FOM experiments, the NINNs being coupled are pre-trained using both the top-down and bottom-up strategies described in Sections \ref{sec:top-down} and \ref{sec:bottom-up}, respectively. 
All NINNs employ the depth-3 Pocket U-Net architecture described in Section \ref{sec:ninns}, with three max-pooling stages, $5\times5$ convolutional kernels, and a fixed width of 32 channels throughout the encoder, bottleneck, and decoder. The convolutional blocks use identity activation functions and no normalization, and a final $1\times1$ convolution maps the decoder output to the scalar nodal solution field. While the same underlying architecture is used in all experiments, the network inputs differ depending on the application, as described below.  The choice of reference solution, evaluation grid, and any additional restrictions on the error calculation depend on the experiment and are specified in the corresponding subsections.


\subsection{Monolithic NINN training at $\Pe_L=10^6$} \label{sec:results_mono}

We first examine whether a NINN can be trained monolithically for the steady advection-diffusion problem \eqref{eq:model} at $Pe_L=10^6$. This experiment is motivated by the PINN-based Schwarz study of Snyder et al. \cite{snyder2023schwarz}, in which monolithic PINN training was unsuccessful at high P\'{e}clet numbers and domain decomposition was introduced, in part, to facilitate PINN training. In particular, Snyder et al. reported that neither the monolithic PINN nor the PINN–PINN Schwarz formulation was successful for P\'{e}clet numbers above 150, whereas PINN–FOM Schwarz coupling enabled a convergent training iteration at $Pe_L=10^6$. These results are consistent with previous studies reporting difficulties in training standard PINNs for advection-dominated problems at high P\'{e}clet numbers.

Here, we investigate whether the NINN formulation can overcome the training difficulty described above without domain decomposition. We consider the monolithic problem \eqref{eq:model} with $f=1$, $~\beta=(1,0)$, and $\nu=10^{-6}$, corresponding to $Pe_L=10^6$. 
With this choice of parameters, the solution exhibits a sharp outflow boundary layer at $x=1$, as well as characteristic boundary layers at $y=0$ and $y=1$.
The NINN and FOM employ the same finite difference discretization: diffusion is approximated using second-order centered differences, while advection is discretized using a second-order upwind scheme with a first-order closure at the first interior node adjacent to the inflow boundary. The FOM solves the resulting discrete system directly, whereas the NINN is trained by minimizing the corresponding discrete residual, as described in Section \ref{sec:ninns}.
The network takes the discretized source term $f_h$ as input and outputs the corresponding nodal solution field. Homogeneous Dirichlet boundary conditions are imposed strongly using the SDBC formulation, as discussed in Section \ref{sec:ninns}. 

For the monolithic experiments in this section, solution accuracy is quantified using a relative discrete $L^2$ error 
\begin{equation} \label{eq:err_steady}
 E_h := \frac{\|~u_h-~u_{\mathrm{ref}}\|_h} {\|~u_{\mathrm{ref}}\|_h}, 
 \end{equation}
where $~u_h$ and $~u_{\mathrm{ref}}$ denote the discrete solution and the reference solution, respectively, evaluated on a common grid, and $\|\cdot\|_h$ is the discrete $L^2$ norm. 
In \eqref{eq:err_steady}, the reference solution $u_{\mathrm{ref}}$ is the analytical solution of \eqref{eq:model}. This solution is obtained using a Fourier sine-series expansion in the $y$-direction, which reduces the 2D advection-diffusion equation to a sequence of 1D boundary value problems for the Fourier coefficients. The evaluation grids used for the individual comparisons are specified below.

To assess the sensitivity of monolithic NINN training to mesh resolution and grading, we consider three mesh resolutions, $33\times33$, $65\times65$, and $129\times129$, referred to as ``coarse", ``medium", and ``fine", respectively. At each resolution, 
we consider two standard forms of layer-adapted meshes.  Hyperbolic-tangent meshes \cite{ThompsonWarsiMastin1985} (referred to as ``$\tanh$" meshes) use a smooth coordinate transformation to cluster nodes near the boundary layer regions, with a grading parameter $s$ controlling the degree of clustering, so that increasing $s$ produces stronger node clustering toward the outflow and characteristic layers.  Herein, we consider $s=1,2,$ and $4$. Shishkin meshes \cite{LinssStynes2001} instead employ a piecewise-uniform construction in which transition points are chosen according to the asymptotic boundary layer scales. For the present problem, letting $N$ denote the number of mesh intervals in each coordinate direction, the Shishkin transition parameters scale as $\tau_x=O(\varepsilon\ln N)$ for the outflow layer and $\tau_y=O(\sqrt{\varepsilon}\ln N)$ for the characteristic layers. Each NINN is trained for 2000 Adam iterations.  

\begin{figure}[ht!]
\centering
\includegraphics[width=\textwidth]{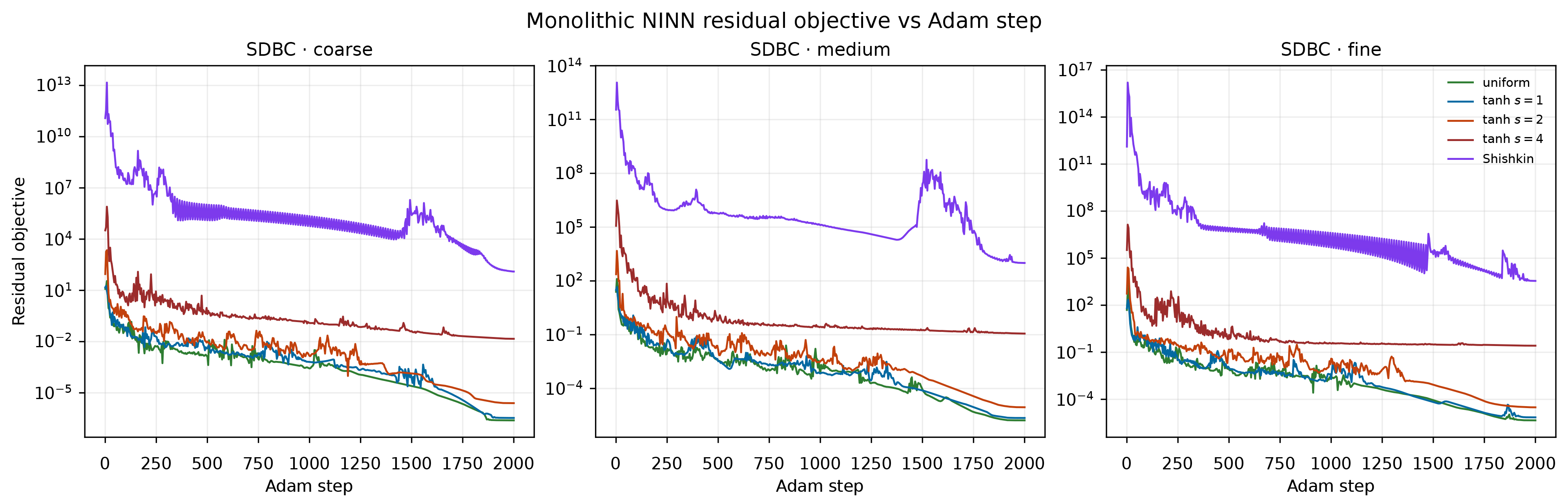}
\caption{NINN loss versus Adam iteration at $Pe_L=10^6$.
Columns
show coarse $(33 \times 33)$, medium $(65\times 65)$, and fine $(129 \times 129)$ meshes.
}
\label{fig:ninn-adam-loss}
\end{figure}

Figure \ref{fig:ninn-adam-loss} shows the NINN training loss as a function of the Adam iteration. On the uniform and mildly graded tanh meshes ($s\leq2$), the residual decreases successfully and the trained NINN closely reproduces the corresponding FOM solution. In contrast, training deteriorates on the strongly graded $s=4$ and Shishkin meshes, with the NINN approaching a near-zero solution in the most strongly graded cases. Although these meshes provide increased resolution in the boundary layer regions, the large variation in local mesh spacing leads to correspondingly large variations in the coefficients of the discrete operator. As a result, the unscaled discrete residual becomes poorly conditioned for optimization, so that improved spatial resolution does not necessarily translate into improved NINN trainability. 
 Importantly, the NINN does not exhibit the same loss of accuracy that is exhibited by a corresponding PINN at $Pe_L = 10^6$ \cite{snyder2023schwarz}. This behavior is examined systematically over a range of P\'{e}clet numbers in Table \ref{tab:Pe_sweep}, and discussed below.\\

\textit{Remark 2.} The reader may wonder whether the NINN results may change if a nonlinear activation function is used within the NINN architecture.  We have observed that choosing the identity activation function or a typical nonlinear activation function has little impact on the accuracy of the NINN solution. In particular, when we replace the identity activations in the convolutional blocks with the tanh activation function for the NINN arhitecture, the $L^2$ relative error for the $Pe_L=10^6$ case changes from $1.50\times 10^{-4}$ to $1.54\times 10^{-4}$.\\  
 

\begin{figure}[ht!]
\centering
\includegraphics[width=\textwidth]{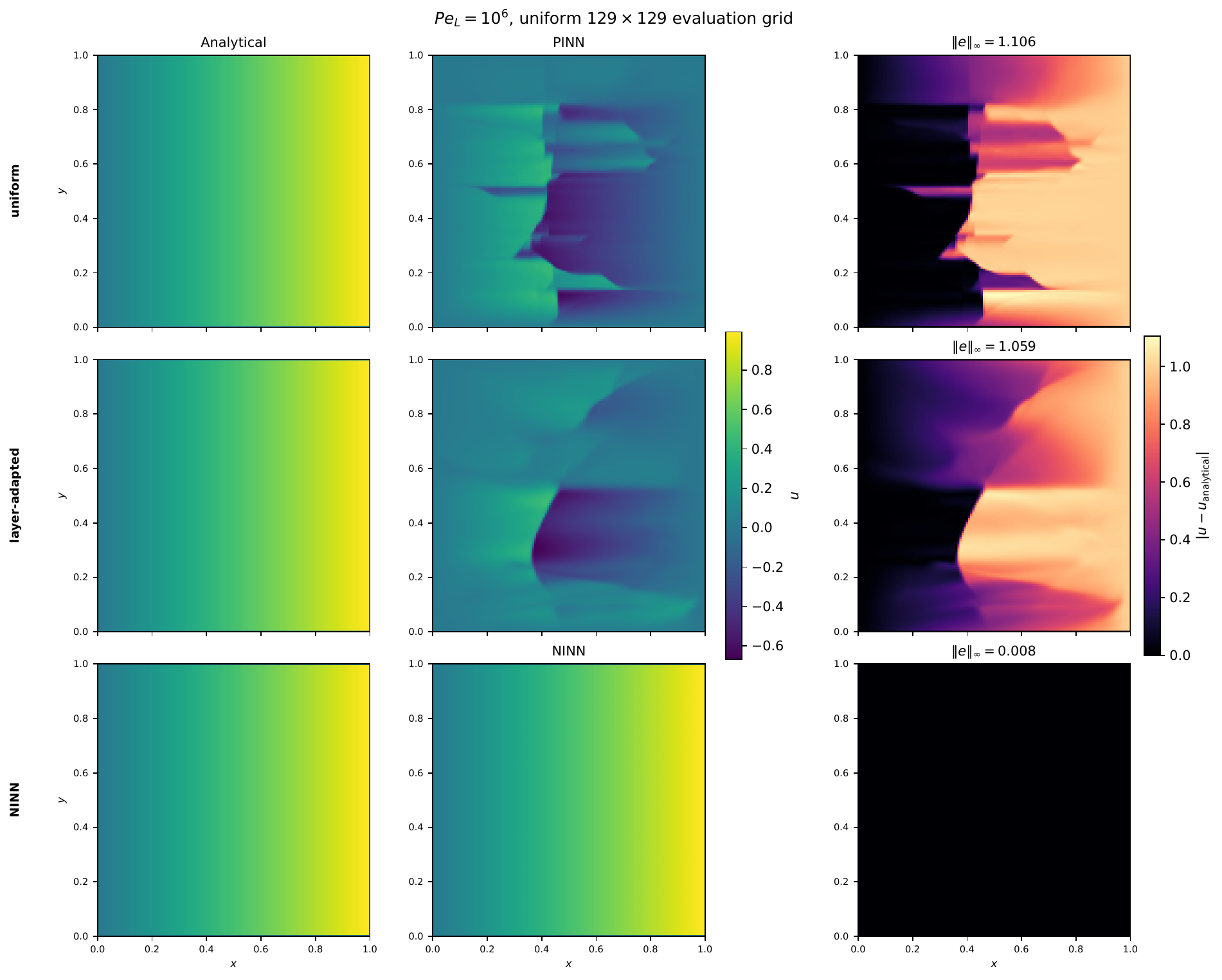}
\caption{Comparison of the analytical solution, coordinate MLP PINN solutions, and NINN solution for $Pe_L=10^6$, evaluated on a uniform $129\times129$ grid. The first two rows show the coordinate MLP PINN results obtained using uniform and layer-adapted interior collocation points, respectively, while the third row shows the NINN result. The right column shows the corresponding pointwise absolute errors with respect to the analytical solution.}
\label{fig:pinn-fields}
\end{figure}

For comparison, we also train a conventional coordinate PINN for the same problem. The PINN is represented by an MLP network with four hidden layers of width 64 and $\tanh$ activation functions. The PDE residual is evaluated in strong form using automatic differentiation, and the homogeneous Dirichlet boundary conditions are imposed strongly as described in Section \ref{sec:ninns} and \cite{snyder2023schwarz}. The PINN is trained for 8000 Adam iterations using 4096 interior collocation points. We consider both uniformly distributed collocation points and a layer-adapted distribution that concentrates points near the outflow and characteristic boundary layers. As shown in Figure \ref{fig:pinn-fields}, neither collocation strategy successfully recovers the $Pe_L=10^6$ solution. The relative $L^2$ errors \eqref{eq:err_steady} with respect to the analytical solution are $\mathcal{O}(1)$ for the uniform and layer-adapted collocation schemes, respectively.  In contrast, the NINN closely reproduces the analytical solution at $Pe_L=10^6$, as shown in the third row of Figure \ref{fig:pinn-fields}.

To determine whether the poor PINN performance can be attributed primarily to the MLP architecture, we additionally train a PINN using the same Pocket U-Net architecture as the NINN, while retaining the strong form PINN residual. We compare the coordinate MLP and Pocket U-Net PINNs for $Pe_L=10,10^2,\ldots,10^6$ on a uniform $33\times33$ evaluation grid. The coordinate PINN uses 4096 Sobol interior collocation points and 8000 Adam iterations, whereas the Pocket U-Net PINN uses 2048 Sobol points and $10^4$ Adam iterations; both are subsequently given a budget of 200 L-BFGS iterations. Because the collocation counts and optimization budgets differ, this experiment is not intended as a controlled comparison of the two PINN architectures; rather, it tests whether the high $Pe_L$ training difficulty persists when the Pocket U-Net architecture is used.

\begin{table}[ht!]
    \centering
    \caption{Relative $L^2$ errors \eqref{eq:err_steady} with respect to the analytical solution for the coordinate MLP PINN, Pocket U-Net PINN, and Pocket U-Net NINN as a function of the P\'{e}clet number $Pe_L$. All errors are evaluated on a uniform $33\times33$ grid.}
    \label{tab:Pe_sweep}
    \begin{tabular}{c|ccc}
        \hline
        $Pe_L$ 
        & MLP PINN 
        & Pocket U-Net PINN 
        & Pocket U-Net NINN \\
        \hline
        $10$     & $5.63\times10^{-4}$ & $1.83\times10^{-3}$ & $9.68\times 10^{-3}$ \\
        $10^2$   & $8.32\times 10^{-2}$& $8.69\times 10^{-1}$ & $4.5\times 10^{-2}$ \\
        $10^3$   & $9.93\times 10^{-1}$           & $9.99\times 10^{-1}$   & $1.11\times 10^{-2}$ \\
        $10^4$   & 1.00      & 1.024      & $8.78\times 10^{-3}$ \\
        $10^5$   & 1.00      & 1.016      & $1.00\times 10^{-3}$ \\
        $10^6$   & 1.00      & 1.015      & $1.50\times 10^{-4}$ \\
        \hline
    \end{tabular}
\end{table}

Table \ref{tab:Pe_sweep} reports the relative $L^2$ errors for the coordinate MLP PINN, Pocket U-Net PINN, and Pocket U-Net NINN over the range of P\'{e}clet numbers considered. At $Pe_L=10$, all three approaches accurately approximate the analytical solution, with both PINNs achieving smaller errors than the NINN. As \(Pe_L\) increases, however, the accuracy of both PINN formulations deteriorates rapidly. The deterioration occurs earlier for the Pocket U-Net PINN, but for $Pe_L\geq10^3$, both PINNs have $O(1)$ relative errors. Thus, although the PINN architecture affects the onset of the loss of accuracy, changing the network architecture alone is not sufficient to overcome the difficulty of training PINNs in the strongly advection-dominated regime.

The NINN exhibits a markedly different trend. Although it is less accurate than either PINN at \(Pe_L=10\), its relative error decreases monotonically for $Pe_L>10$, reaching $1.50\times10^{-4}$ at \(Pe_L=10^6\). The reason for the greater accuracy of the PINNs at $Pe_L=10$, as well as the subsequent improvement in NINN accuracy with increasing $Pe_L$, is not clear from the present experiments. Nevertheless, the contrasting behavior of the Pocket U-Net PINN and NINN indicates that the high $Pe_L$ performance of the NINN cannot be attributed solely to its network architecture and may instead be related to its use of a prescribed finite difference discretization in the training objective rather than the strong-form PDE residual used by the PINNs. A more systematic study would be needed to determine the mechanism responsible for this behavior. \\

Taken together, the results described above indicate that the successful training of the
NINN at $Pe_L$ up to $10^6$ cannot be attributed solely to its use of the
Pocket U-Net architecture. The substantially different behavior of the NINN
be related to its use of a prescribed finite difference
discretization in the training objective, rather than the strong form PDE
residual used by the PINNs. 
A more systematic
study would be needed to determine the mechanism responsible for the observed
improvement in NINN accuracy with increasing $Pe_L$. Finally, we note that
our PINN results are consistent with previous
studies reporting substantial training difficulties for standard PINNs in
advection-dominated regimes, e.g., \cite{snyder2023schwarz}.

The study described above leads to the conclusion that,  unlike the PINN-based approach of Snyder et al.~\cite{snyder2023schwarz}, there is no need to use Schwarz domain decomposition to facilitate NINN training. Instead, in the following section, Schwarz serves as a coupling mechanism between a pre-trained NINN and a FOM, with the NINN parameters held fixed throughout the online Schwarz iteration.\\

\noindent \textit{Remark 3.} Recall that the PINNs considered here omit a data loss term and are trained solely from the governing equations and boundary conditions. The reader may therefore wonder whether the performance of the PINNs could be improved by incorporating a data-loss term. Such an approach, however, makes the training supervised, requiring reference solution data that may be expensive to generate or unavailable for the problem of interest. In \cite{snyder2023schwarz}, Snyder et al. considered PINNs augmented with a data-loss term and nevertheless observed difficulties in the advection-dominated regime. In contrast, the NINNs considered here achieve remarkably high accuracy in a substantially more advection-dominated regime using completely unsupervised training, without requiring any reference solution data. \\ 

\subsection{Pre-trained NINN-FOM coupling}
\label{sec:results_steady}

We next consider the use of the overlapping Schwarz method to couple a pre-trained NINN with a FOM for the steady advection-diffusion problem \eqref{eq:model} at $Pe_L=10^6$ posed on the subdomains shown in Figure \ref{fig:dd} with $\gamma_L = 0.95$ and $\gamma_R = 0.65$.  A pre-trained NINN is employed on $\Omega_L$, while the finite difference FOM is employed on 
$\Omega_R$, as $\Omega_R$ contains a sharp outflow layer at $x=1$. Both local models use uniform $33\times33$ meshes. The NINN parameters are held fixed throughout the online Schwarz iteration, as described in Section \ref{sec:ninn-fom_schwarz}.  

To train the left subdomain NINN over a range of source terms and interface boundary conditions, we consider the source family
\begin{equation} f_q(~x)=c_{q,0}+\sum_{\ell=1}^{3}c_{q,\ell}\sin(\pi x)\sin(\ell\pi y), 
\end{equation}
where $~x := (x,y)$,  $c_{q,0}\sim\mathcal{U}(0.75,1.25)$ and $c_{q,\ell}\sim\mathcal{U}(-0.5,0.5)$ for $\ell=1,2,3$, are sampled independently, with $\mathcal{U}(a,b)$ denoting the uniform distribution on $[a,b]$. A total of 256 source functions are generated for training. For testing, we use the constant source $f=1$, which is explicitly excluded from training, together with 16 additional independently sampled sources from (5.2), also disjoint from the training set.

For each training source, Dirichlet boundary data on the artificial interface $\Gamma_L$ are generated using either the top-down or bottom-up strategy introduced in Section~\ref{sec:ninn_training}. In the top-down approach, a FOM-FOM Schwarz problem is solved for each of the 256 training sources, and the converged Dirichlet data on $\Gamma_L$ are retained for NINN training. For the present problem, each FOM-FOM Schwarz solve converges after two iterations, yielding one nonzero interface trace per source and hence 256 top-down training pairs.

In the bottom-up approach, the 256 interface traces are instead generated synthetically according to \eqref{eq:sine}, with $M=8$. Specifically, the coefficients are sampled independently as $\xi_{q,m}\sim\mathcal{N}(0,1)$, where $\mathcal{N}(0,1)$ denotes the standard normal distribution, and each resulting trace is normalized and rescaled to have a randomly sampled peak amplitude $\rho_q\sim\mathcal{U}(0.25,1.25)$. Thus, the bottom-up approach uses the same family of 256 source terms as the top-down approach but replaces the FOM-generated interface data by independently generated synthetic traces.

For each training strategy, the NINN is trained over the 256 source/interface pairs using 8000 Adam updates. The learning rate is cosine-annealed from $10^{-3}$ to zero, with $\epsilon=10^{-7}$, zero weight decay, and gradient clipping at $10^{-2}$. To account for variability due to network initialization, three independent training runs are performed using different random initializations of the NINN parameters. For a consistent comparison between the two training strategies, each top-down NINN and its corresponding bottom-up NINN are initialized with the same parameter values and subsequently trained independently.


Following training, the six resulting pre-trained NINNs -- three for each training strategy -- are then deployed in the hybrid Schwarz iteration for the held-out constant source $f=1$ and the 16 additional source functions excluded from training. 
The absolute and relative tolerances in the Schwarz convergence criterion \eqref{eq:schwarz_conv} are set to $\tau_a=\tau_r=10^{-6}$ for all simulations.  This gives $3\times17=51$ hybrid solves for each training strategy, or 102 coupled NINN-FOM solves in total. Solution accuracy is measured relative to the corresponding FOM-FOM Schwarz solution using the global composite solution described above. Specifically, all converged solutions are evaluated on a uniform
$129\times65$ diagnostic grid $P$, and the relative error is computed as
\begin{equation} \label{eq:schwarz_err}
    E_h^{\mathrm{Schwarz}}
    =
    \frac{\|~u_h^{\mathrm{NINN-FOM}}-~u_h^{\mathrm{FOM-FOM}}\|_P}
    {\|~u_h^{\mathrm{FOM-FOM}}\|_P},
\end{equation}
where
\begin{equation}
    \|~v\|_P^2 := \sum_{p\in P} w_p v(p)^2
\end{equation}
is an area-weighted discrete norm, $~v$ denotes a discrete field evaluated
on $P$, and $w_p$ is the corresponding nodal area weight. The weights
$w_p$ correspond to the composite trapezoidal rule on the uniform
diagnostic grid. Denoting the uniform grid spacings in the $x$- and
$y$-directions by $\Delta x$ and $\Delta y$, respectively,
$w_p=\Delta x\,\Delta y$ at interior grid points, with the weights reduced
by a factor of two along boundary edges and by a factor of four at
corners.
In \eqref{eq:schwarz_err}, $~u_h^{\mathrm{FOM-FOM}}$ and
$~u_h^{\mathrm{NINN-FOM}}$ denote the global composite solutions obtained
from the converged FOM-FOM and NINN-FOM Schwarz iterations, respectively.
In both cases, the composite solution is formed using a partition cut
at $x_c=0.8$, with the left subdomain solution used for $x\leq x_c$ and the
right subdomain solution used for $x>x_c$.  The diagnostic grid is used only for evaluating this error and is distinct from the $33\times33$ local meshes used in the Schwarz solves.

\begin{table}[ht!]
\centering
\caption{Relative errors for the pre-trained NINN-FOM Schwarz coupling experiments.
For $f= 1$, the reported range is over the three independently pre-trained
NINNs for each training strategy. For the 16 held-out sources, the range is
over the 48 runs obtained from the 16 sources and three independently trained
NINNs for each training strategy.}
\label{tab:steady_coupling_errors}
\begin{tabular}{llccc}
\toprule
Test set & Training strategy & No. of runs & Median $E_h^{\text{Schwarz}}$ & Range of $E_h^{\text{Schwarz}}$ \\
\midrule
$f = 1$          & Top-down  & 3  & 0.062\% & 0.035--0.079\% \\
$f = 1$          & Bottom-up & 3  & 0.075\% & 0.061--0.101\% \\
16 held-out sources   & Top-down  & 48 & 0.050\% & 0.027--0.221\% \\
16 held-out sources   & Bottom-up & 48 & 0.053\% & 0.037--0.284\% \\
\bottomrule
\end{tabular}
\end{table}

\begin{figure}[ht!]
\centering
\includegraphics[width=\textwidth]{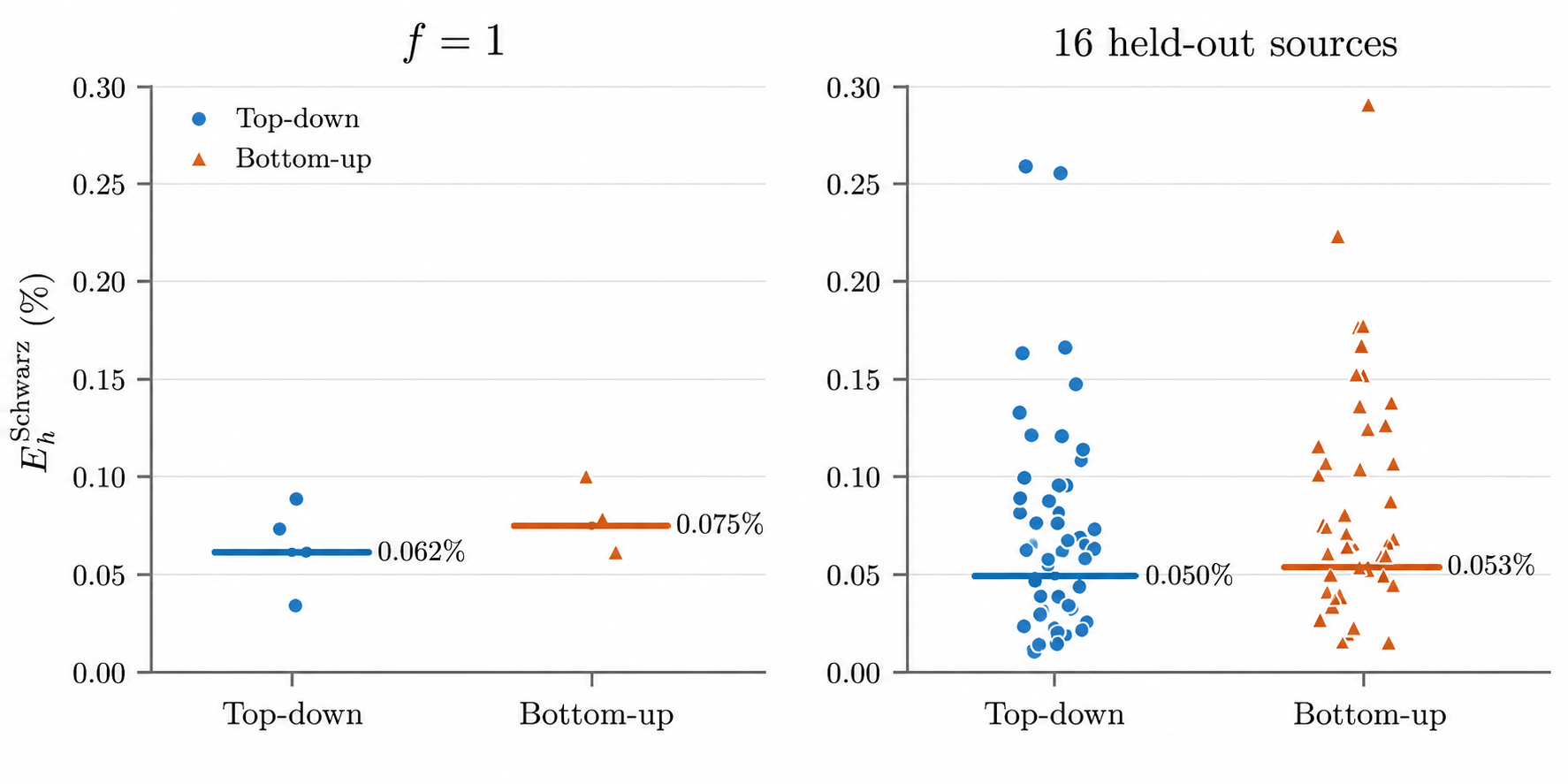}
\caption{Relative Schwarz coupling error $E_h^{\mathrm{Schwarz}}$ for the pre-trained NINN-FOM coupling experiments at $Pe_L=10^6$. \textit{Left:} held-out constant source $f=1$. Each point corresponds to one of the three independently trained NINNs for the indicated training strategy. \textit{Right:} 16 held-out sources. Each point corresponds to one NINN-FOM solve for a particular held-out source and trained NINN, giving 48 points per training strategy. Horizontal bars indicate the median error over the points shown for each training strategy.}
\label{fig:pre-trained-error}
\end{figure}

Table \ref{tab:steady_coupling_errors} and Figure \ref{fig:pre-trained-error} summarize the Schwarz coupling errors for the two training strategies. Both the top-down and bottom-up NINNs yield hybrid solutions in close agreement with the corresponding FOM-FOM Schwarz solutions. For the held-out constant source $f=1$, the median errors are $0.062\%$ and $0.075\%$ for the top-down and bottom-up approaches, respectively, with little variation among the three independently trained NINNs. Across the 16 additional held-out sources, the corresponding median errors are $0.050\%$ and $0.053\%$, with similar distributions of errors. Although a small number of source/NINN combinations exhibit larger errors, all errors remain below $0.3\%$. These results indicate that, for the problems considered here, synthetically generated bottom-up interface traces provide coupling accuracy comparable to that obtained using FOM-generated top-down traces, while avoiding the need for coupled FOM-FOM simulations during the offline data generation stage.


\begin{figure}[ht!]
\centering
\includegraphics[width=\textwidth]{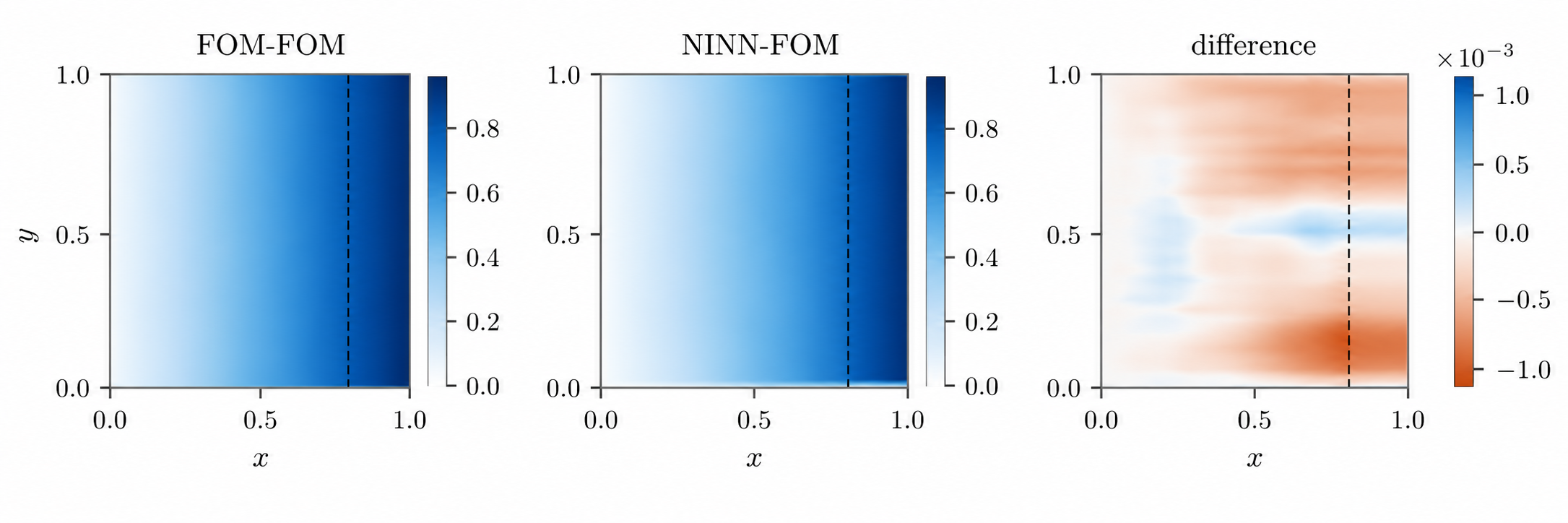}
\caption{FOM-FOM Schwarz composite, NINN-FOM Schwarz composite, and their signed pointwise difference, $(\mathbf{u}_h^{\mathrm{NINN-FOM}}-\mathbf{u}_h^{\mathrm{FOM-FOM}})$, for $f= 1$ at $Pe_L=10^6$, evaluated on the diagnostic grid $P$. The NINN-FOM solution uses the top-down NINN from the first paired initialization. The dashed line denotes the partition cut $x_c=0.8$, at which each composite switches from the left to the right subdomain solution.}
\label{fig:pre-trained-fields}
\end{figure}

Finally, Figure \ref{fig:pre-trained-fields} provides a representative comparison of the FOM-FOM and NINN-FOM composite solutions for the held-out source $f=1$. The two composite fields are visually indistinguishable at the scale of the solution, while their signed pointwise difference remains $\mathcal{O}(10^{-3})$. The maximum absolute pointwise difference is $1.1\times10^{-3}$ and occurs near the partition cut $x_c=0.8$. Thus, replacing the left subdomain FOM with the pre-trained NINN introduces only a small perturbation to the converged Schwarz composite solution.

We conclude by comparing the computational cost of the NINN-FOM and FOM-FOM Schwarz couplings. As summarized in Table \ref{tab:steady_coupling_cost}, the FOM-FOM Schwarz coupling converges in two iterations, with wall-clock times ranging from $0.013$ to $0.021$ s, whereas the NINN-FOM Schwarz method requires three to seven iterations, with wall-clock times ranging from 
$0.057$ to $0.235$ s. The increase in Schwarz iterations for the hybrid coupling is expected: because the NINN provides an approximate representation of the corresponding FOM solution operator, additional Schwarz iterations may be required before the exchanged traces satisfy the same stopping test. Similar behavior has been observed previously for Schwarz coupling involving projection-based reduced-order models \cite{Wentland2025}.
In addition, offline training of each NINN requires $158$–$171$ s. 

The reader can observe from Table \ref{tab:steady_coupling_cost} that the present NINN-FOM implementation does not provide a computational speedup relative to the FOM-FOM baseline for this small problem. The increase in online wall-clock time is larger than would be expected from the Schwarz iteration count alone. This is likely due to the relative cost of the local solvers: for the present \(33\times33\) test problem, each FOM subdomain solve is extremely inexpensive, as the finite difference FOM implementation used here has been highly optimized.  
In contrast, a NINN evaluation incurs the fixed cost of multiple convolutional, pooling, and decoder operations. Consequently, neural network inference overhead can dominate at this small problem size. This result should therefore not be interpreted as evidence that NINN-based local models are intrinsically more expensive than FOMs; however, further studies would be required to study whether NINNs are more cost-effective on larger benchmarks. The objective of the present experiments is instead to assess the accuracy and feasibility of coupling a pre-trained NINN with a FOM; reducing the computational cost of the NINN local solve will be studied in future work.

\begin{table}[ht!]
    \centering
    \caption{Computational cost of the FOM-FOM and NINN-FOM 
    Schwarz coupling experiments.}
    \label{tab:steady_coupling_cost}
    \begin{tabular}{lccc}
        \toprule
        Method & Schwarz iterations & Online time (s) & Offline training time (s) \\
        \midrule
        FOM-FOM  & 2   & 0.013--0.021 & -- \\
        NINN-FOM & 3--7 & 0.057--0.235 & 158--171 \\
        \bottomrule
    \end{tabular}
\end{table}

\section{Conclusions and future work}  \label{sec:conc}

In this work, we investigated the use of the overlapping Schwarz alternating
method as a framework for coupling pre-trained NINNs with FOMs. More broadly, the work is
motivated by the need for systematic approaches for incorporating data-driven
models into traditional modeling and simulation workflows. The Schwarz
framework provides a natural mechanism for doing so by allowing different
models to be deployed in different regions of the computational domain. The
underlying idea is to employ a data-driven surrogate in subdomains where the
solution is comparatively easier to approximate, while retaining a FOM in
regions containing more challenging solution features. In this way, the
accuracy and robustness of the FOM can be retained where they are most needed,
while the data-driven model can potentially reduce the computational cost
elsewhere. In the present work, we investigate this approach by training a NINN offline as a subdomain-local surrogate and subsequently coupling it to a neighboring FOM using the overlapping Schwarz alternating method. Dirichlet data are exchanged between the NINN and FOM subdomains during the Schwarz iteration, while the NINN parameters remain fixed. We
demonstrated this framework for a 2D steady
advection-diffusion problem in a highly advection-dominated regime with
$Pe_L=10^6$.

We first examined the ability of a monolithic NINN to approximate the
high P\'eclet-number solution without domain decomposition. A NINN with
strongly enforced Dirichlet boundary conditions was successfully trained at
$Pe_L=10^6$, whereas PINNs based on both a conventional coordinate MLP and
the same Pocket U-Net architecture employed by the NINN exhibited substantial
loss of accuracy as the P\'eclet number increased. The latter comparison
indicates that the improved performance of the NINN cannot be attributed
solely to the choice of NN architecture. Rather, an important
distinction is that the NINN training objective incorporates a prescribed
discretization of the governing PDE, in contrast to the strong form PDE
residual employed by the PINN. Moreover, the NINN achieves this accuracy in a
fully unsupervised setting, without requiring reference solution data for
training.

We then demonstrated that a pre-trained, subdomain-local NINN can be
successfully coupled with a neighboring FOM using the overlapping Schwarz
method. The resulting NINN-FOM solutions were in close agreement with the
corresponding FOM-FOM Schwarz solutions for both the reserved constant source
and a collection of source functions excluded from training. We considered
two strategies for generating the interface data used during offline NINN
training: top-down and bottom-up training. 
Both strategies produced accurate hybrid NINN-FOM solutions,
suggesting that sufficiently rich synthetic interface data can enable
effective deployment of a pre-trained NINN without requiring FOM-FOM
Schwarz simulations for training.

The present work has exposed several natural directions for future research. A central objective of the hybrid
framework is to reduce computational cost by replacing expensive FOM solves
with data-driven surrogates in selected subdomains while retaining FOMs in
regions where they are needed to accurately resolve more challenging solution
features. The present NINN--FOM implementation does not yet provide a computational speedup relative to the corresponding FOM--FOM solve. Achieving such a speedup will require more efficient NINN inference and a reduction in the overall cost of the hybrid Schwarz iteration.  In addition to reducing the cost of the NINN evaluation itself,
future work will investigate the use of acceleration techniques, such as
Aitken and Anderson acceleration~\cite{sambataro2026relaxation}, to reduce the
number of Schwarz iterations required for convergence.

Beyond computational efficiency, there are several natural extensions of the
present work. We plan to extend the NINN-FOM Schwarz framework to
time-dependent and nonlinear problems, as well as to more complex geometries
and three-dimensional problems. Another important direction is removing the
current restriction to structured finite difference grids by developing NINNs
for unstructured meshes and other spatial discretizations, including finite
element methods \cite{celaya2025dgninn}. We also plan to consider more general Schwarz configurations,
including multiple subdomains, NINN-NINN coupling, heterogeneous
discretizations, and nonconforming interfaces. Finally, the present numerical
experiments consider a relatively limited family of forcing functions and a
fixed set of physical parameters. Future studies will examine how well
offline-trained NINNs generalize as the forcing, boundary and interface data,
and physical parameters are varied, which will be necessary for applying the
approach to realistic multiphysics and multiscale problems.

\section*{Acknowledgements} \label{sec:acknowl}

\begin{sloppypar}
Support for this work was received through Sandia National Laboratories' Laboratory Directed Research and Development (LDRD) program and through the U.S. Department of Energy, Office of Science, Office of Advanced Scientific Computing Research, Mathematical Multifaceted Integrated Capability Centers (MMICCs) program, under Field Work Proposal 22025291 and the Multifaceted Mathematics for Predictive Digital Twins (M2dt) project. Additionally, the writing of this manuscript was funded in part by Irina Tezaur’s Presidential Early Career Award for Scientists and Engineers (PECASE).
\end{sloppypar}

Sandia National Laboratories is a multi-mission laboratory managed and operated by National Technology and Engineering Solutions of Sandia, LLC., a wholly owned subsidiary of Honeywell International, Inc., for the U.S. Department of Energy’s National Nuclear Security Administration under contract DE-NA0003525.

\raggedbottom
\bibliographystyle{siam}
\bibliography{GeorgeChumbipuma}

\end{document}